\documentclass[conference]{IEEEtran}
\IEEEoverridecommandlockouts
\usepackage{cite}
\usepackage{amsmath,amssymb,amsfonts}
\usepackage{algorithmic}
\usepackage{graphicx}
\usepackage{textcomp}
\usepackage{xcolor}

\usepackage{times}
\usepackage{epsfig}

\usepackage{bm}  
\usepackage{multirow}  
\usepackage{wrapfig}  
\usepackage[
            left=0.75in,right=0.75in,top=1in,bottom=0.75in,%
            ]{geometry}  

\DeclareMathOperator*{\argmin}{arg\,min}
\newcommand{\id}{\mathrm{id}}

\def\BibTeX{{\rm B\kern-.05em{\sc i\kern-.025em b}\kern-.08em
    T\kern-.1667em\lower.7ex\hbox{E}\kern-.125emX}}
\begin{document}

\title{Test-Time Training for Deformable \\ Multi-Scale Image Registration

\thanks{Corresponding author: Xiaohui Xie, University of California, Irvine (xhx@ics.uci.edu).}
}


\author{Wentao Zhu$^{1}$ \hspace{0.5cm} Yufang Huang$^{2}$ \hspace{0.5cm} Daguang Xu$^{3}$ \hspace{0.5cm} Zhen Qian$^{4}$ \hspace{0.5cm} Wei Fan$^{4}$ \hspace{0.5cm} Xiaohui Xie$^{5}$ \\
$^1$Kuaishou Technology \hspace{0.5cm} $^2$Cornell University \hspace{0.5cm} $^3$NVIDIA \hspace{0.5cm} $^4$Tencent \hspace{0.5cm} $^5$University of California, Irvine }



\maketitle

\begin{abstract}
Registration is a fundamental task in medical robotics and is often a crucial step for many downstream tasks such as motion analysis, intra-operative tracking and image segmentation. Popular registration methods such as ANTs and NiftyReg optimize objective functions for each pair of images from scratch, which are time-consuming for 3D and sequential images with complex deformations. Recently, deep learning-based registration approaches such as VoxelMorph have been emerging and achieve competitive performance. In this work, we construct a test-time training for deep deformable image registration to improve the generalization ability of conventional learning-based registration model. We design multi-scale deep networks to consecutively model the residual deformations, which is effective for high variational deformations. Extensive experiments validate the effectiveness of multi-scale deep registration with test-time training based on Dice coefficient for image segmentation and mean square error (MSE), normalized local cross-correlation (NLCC) for tissue dense tracking tasks.  
\end{abstract}


\section{Introduction}
\label{sec:intro}
Image registration tries to establish the correspondence between organs, tissues, landmarks, edges, or surfaces in different images and it is critical to many clinical tasks such as tumor growth monitoring and surgical robotics~\cite{haskins2019deep}. Manual image registration is time-consuming, laborious and lacks reproducibility which causes clinical disadvantage potentially. Thus, automated registration is desired in many clinical practices. Generally, registration can be necessary to analyze motion from videos, auto-segment organs given atlases, and align a pair of images from different modalities, acquired at different times, from different viewpoints or even from different patients. Thus designing a robust image registration can be challenging due to the high variability of scenarios.

Traditional registration methods estimate the registration fields by optimizing certain objective functions. Such registration field can be modeled in several ways, e.g. elastic-type models~\cite{bajcsy1989multiresolution,shen2002hammer}, free-form deformation~\cite{rueckert1999nonrigid}, Demons~\cite{thirion1998image}, and statistical parametric mapping~\cite{ashburner2000voxel}. Beyond the deformation model, diffeomorphic transformations preserve topology and many methods adopt them such as large deformation diffeomorphic metric mapping~\cite{beg2005computing}, symmetric image normalization (SyN)~\cite{avants2008symmetric} and diffeomorphic anatomical registration using exponential Lie algebra~\cite{ashburner2007fast}. One limitation of these methods is that the optimization can be computationally expensive especially for 3D images.

Deep learning-based registration methods have recently been emerging as a practicable alternative to the conventional registrations~\cite{rohe2017svf,sokooti2017nonrigid,yang2017quicksilver,wu2013unsupervised,blendowski2019combining,hu2016efficient,liao2017artificial,ma2017multimodal,miao2018dilated,krebs2017robust}. These methods employ sparse or weak label of registration field, or conduct supervised learning~\cite{chee2018airnet,rohe2017svf,cao2017deformable,sokooti2017nonrigid,uzunova2017training,hur2019iterative} purely based on registration field ground truth. Facilitated by a spatial transformer network~\cite{jaderberg2015spatial}, recent unsupervised deep learning-based registrations~\cite{mok2020large,neylon2017neural,li2018non,sheikhjafari2018unsupervised,godard2017unsupervised,fan2018adversarial,jiang2018cnn,de2019deep,zhao2019recursive}, such as VoxelMorph~\cite{balakrishnan2018unsupervised,de2017end}, have been explored because of annotation free in the training. VoxelMorph is further extended to diffeomorphic transformation and Bayesian framework~\cite{dalca2018unsupervised}. Adversarial similarity network adds an extra discriminator to model the appearance loss between warped image and fixed image, and uses adversarial training to improve the unsupervised registration~\cite{zhu2018adversarial,fan2018adversarial,shen2020multi,huang2020cycle}. However, most of these registration methods have comparable accuracy as traditional registrations, although with potential speed advantages.

In this work, we design a novel test-time training for deep deformable image registration framework with multi-scale parameterization of complicated deformations for accurate registration to handle \textit{large deformation and noise} widely existing in medical images. The framework, called self-supervision optimized multi-scale registration as illustrated in Fig.~\ref{fig:framework}, is motivated by the fact that these purely learning-based registrations generally cannot generalize well on noisy images with large and complicated deformations because of domain shift between training data and test data~\cite{sun2019test}. More specifically, we propose a registration optimized in both training and test to tackle the \textit{generalization gap} between training images and test images. Different from unsupervised registration~\cite{balakrishnan2018unsupervised}, registration with test-time training~\cite{sun2019test} further tunes the deep network in each test image pair, which is vital for the success of noisy image registration in Fig.~\ref{fig:muscle} and~\ref{fig:blood}. Inspired by the improvement of multi-scale registration in conventional registration~\cite{curiale2016influence}, we further employ a multi-scale scheme to estimate accurate deformation where we conduct test-time training with a U-Net~\cite{ronneberger2015u} to model the \textit{residual registration field} in each scale.
{\small{
\begin{figure}[t]
\centering
    \includegraphics[width=0.50\textwidth,trim=0 0 0 0,clip]{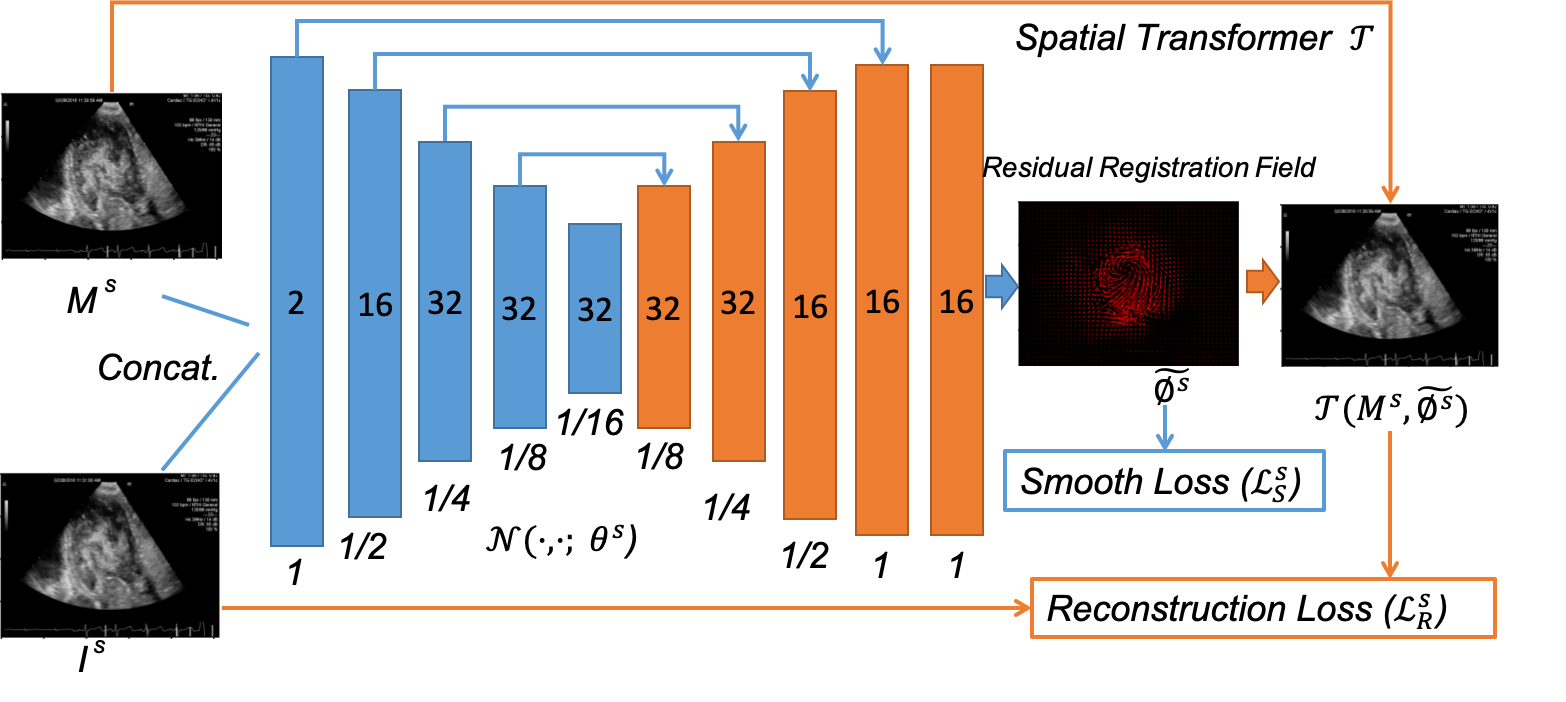}  
\caption{Framework of \textbf{\textit{multi-scale}} deep registration networks with \textbf{\textit{test-time training}} in scale $s$.}
\label{fig:framework}
\end{figure}}}

Our main contributions are as follows. 1) We design a novel test-time training for deep deformable registration to improve generalization ability of learning-based registration. The test-time training with self-supervised optimization for deep registration network yields accurate registration field estimation even for noisy images with large deformations by eliminating the accuracy gap between the estimation on the training set and that on the test set. 2) We design a deep multi-scale registration based on unsupervised learning framework to model the large deformations. The multi-scale strategy estimates the \textit{residual registration field} consecutively, which enforces a coarse to fine
consistency of the morphing, and it provides a sequential optimization pathway which yields much more accurate registration field.  

\section{Method}\label{sec:data_method}

\subsection{Notations and Framework}\label{sec:denot}

Let $\bm{M}$ and $\bm{I}$ be two images defined over an n-dimensional spatial domain $\mathrm{\Omega}\in \mathbb{R}^n$. In this paper, we focus on either $n=2$ for 2D images such as X-ray and ultrasound, or $n=3$ for volumetric images such as computed tomography (CT) or magnetic resonance imaging (MRI). For simplicity, we assume both $\bm{M}$ and $\bm{I}$ are gray-scale containing a single channel. Our goal is to align $\bm{M}$ to $\bm{I}$ through a deformable transformation so that the anatomical features in transformed $\bm{M}$ are aligned to those in $\bm{I}$. To distinguish them, we call $\bm{M}$ and $\bm{I}$ the moving and fixed images, respectively. 

Let $\phi\colon \mathrm{\Omega} \to \mathrm{\Omega}$ be the registration field that maps coordinates of $\bm{I}$ to coordinates of $\bm{M}$. The image registration problem \cite{avants2009advanced,bajcsy1989multiresolution,balakrishnan2019voxelmorph} can be generally formulated as minimizing the anatomical differences between $\bm{M}\circ \phi$ ($\bm{M}$ warped by $\phi$) and $\bm{I}$, regularized by a smoothness constraint on the registration field,  
\begin{equation}\label{eq:loss}
\hat \phi = \argmin_\phi\ \mathcal{L}_R(\bm{I}, \bm{M}\circ \phi) + \lambda \mathcal{L}_S(\phi)
\end{equation}
where $\mathcal{L}_R$ is a reconstruction loss measuring the dissimilarity between two images, and $\mathcal{L}_S$ measures the smoothness of the registration field. In this work, we assume $\phi = \id + \mathbf{u}$ with $\id$ denoting the identify mapping, and model the displacement field $\mathbf{u}$ instead. Let $u_i$ denote the $i$-th coordinate of the displacement field. The smoothness term is taken to be $$
\mathcal{L}_S(\phi) = \sum_{\mathbf{p}\in\mathrm{\Omega}}\sum_{i=1}^n\|\nabla u_i(\mathbf{p})\|^2
$$
throughout this paper with the gradients approximated by displacement differences between neighboring grids in the actual implementation.  

\subsection{Deep Learning-Based Registration}
We model the displacement field through a deep neural net, $\mathbf{u}=\mathcal{N}(\bm{I}, \bm{M}; \bm{\theta})$, receiving $\bm{I}$ and $\bm{M}$ as input and generating $\mathbf{u}$ through a neural net with weight parameters $\bm{\theta}$. Finding the registration field is thus expressed as a learning problem to identify the optimal parameters $\bm{\theta}$ to minimize the loss function shown in Eq.~\eqref{eq:loss}. We employ a U-Net~\cite{ronneberger2015u} $\mathcal{N}(\cdot, \cdot; \bm{\theta})$ with skip connections to estimate the displacement field $\mathbf{u}$. The input to the network is the concatenation of  $\bm{M}$ and $\bm{I}$ across the channel dimension. The U-Net configuration is illustrated in Fig.~\ref{fig:framework}. 

Given a point $\mathbf{p}\in\mathrm{\Omega}$,  the registration field aligns its image feature $\bm{I}(\mathbf{p})$ in the fixed image to the image feature at location $\phi(\mathbf{p})$ of the moving image. Because image values are only defined over a grid of integer points, we linearly interpolate the values of $\bm{M}\circ\phi(\mathbf{p})$ from the grid points neighboring $\phi(\mathbf{p})$ through a spatial transformer network $\mathcal{T }$~\cite{jaderberg2015spatial}, which generates a warped moving image according to the registration field $\bm{I}_R = \mathcal{T}(\bm{M}, \phi)$. This formulation allows gradient calculation and back-propagating errors during learning. 

Many metrics have been proposed to measure the anatomical similarity between fixed image $\bm{I}$ and warped moving image $\bm{I}_R$, including mean-squared error, mutual information, cross-correlation. In this work, we use the negative normalized local cross-correlation as the reconstruction loss, which tends to be more robust for noisy images, although other metrics can be equally applied
\begin{equation}
\begin{aligned}
&\mathcal{L}_{R}(\bm{I}, \bm{M}\circ\phi) =\\ &-\frac{1}{|\mathrm{\Omega}|}\sum_{\bm{p} \in \mathrm{\Omega}} \frac{ \big( \sum_{\bm{q}} {(\bm{I}(\bm{q}) - \overline{\bm{I}(\bm{p})} ) (\bm{I}_R(\bm{q}) - \overline{\bm{I}_R(\bm{p})} )} \big)^2}{\sum_{\bm{q}} (\bm{I}(\bm{q}) - \overline{\bm{I}(\bm{p})} )^2   \sum_{\bm{q}} (\bm{I}_R(\bm{q}) - \overline{\bm{I}_R(\bm{p})} )^2 }, 
\end{aligned}
\label{eq:l_R}
\end{equation}
where $\bm{q}$ is a location within a local neighborhood around $\bm{p}$, and $\overline{\bm{I}(\bm{p})}$ and $\overline{\bm{I}_R(\bm{p})}$ are the local means of the fixed and warped images, respectively.

\subsection{Test-Time Training for Registration}\label{sec:ssr}
Given training and test data, previous neural based registration methods take the standard paradigm of learning the parameters of the neural net by minimizing the loss function Eq.~\eqref{eq:loss} on the training set, and then derive the displacement field on the test set based on the learned parameters. The benefit of this approach is that the inference is fast but coming at the cost of significant performance reduction. The main cause for this is that each pair of medical images has its own unique characteristics and it is difficult for learned models to generalize well on new test images. For medical images with low signal-to-noise ratios, such as ultrasound images,  the performance gap between the training and test sets can be substantial. 

We propose to use self-supervised optimization to improve the generalization accuracy of learning-based registration. Under this paradigm, the network parameters $\bm{\theta}$ is further optimized based on both training and the test image pairs
\begin{align}
    &\bm{\theta}_{\star} = \mathop{\arg \min}_{\bm{\theta}} \mathbb{E}_{(\bm{M}, \bm{I}) \in \mathcal{D}} [{\mathcal{L}_{R}(\bm{I}, \mathcal{T }(\bm{M}, \phi); \bm{\theta}) + \lambda \mathcal{L}_{S}(\phi; \bm{\theta})}],
    \label{eq:sso}
\end{align}
where $\mathcal{D}$ contains both the image pairs in the training set as well as test image pairs, and $\lambda$ is a regularization parameter balancing the trade-off between the reconstruction and smoothness losses.

Our approach aims to seek a middle ground between traditional optimization-based approaches and pure learning-based approaches. The neural network parameterized registration trained with stochastic optimization alleviates some of key challenges in traditional optimization-based approaches such as high cost in optimization, long running time, and poor performance due to local optimality~\cite{kingma2014adam}, while at the same time improves the performance of purely learning-based approach through further fine tuning on test images. 

\subsection{Multi-Scale Neural Registration}\label{sec:msf}
The registration field for medical images often have both large and small-scale displacement vectors throughout the spatial domain. This is most apparent in echocardiogram, where we use image registration to align temporally nearby images to detect tissue or blood flow movements. In order to capture the displacement field at various scales, we propose a multi-scale scheme to parameterize the \textit{residual registration field}. At each scale, a self-supervision optimized registration network uses the concatenation of reconstructed image from last scale and fixed image of current scale as input. A U-Net is used to parameterize the residual registration field of each scale. The final registration field is calculated by fusing the residual registration fields across different scales. 

We choose a sequence of spatial scales $\{S, 2S, \cdots, s, \cdots, 1/2, 1\}$ from coarse to fine to parameterize the neural registration field, with each modeled by a U-Net. We employ resize with different image sizes for different spatial scales. For instance, $\mathcal{N}(\cdot, \cdot; \bm{\theta}^s)$ models the neural registration field $\tilde\phi^s$ at scale $s$. At each scale, we use self-supervised optimization to obtain the parameter $\bm{\theta}^s$ in the corresponding U-Net.

At each scale $s$, the input to the neural registration field $\mathcal{N}(\cdot, \cdot; \bm{\theta}^s)$ is the concatenation of reconstructed image $\bm{M}^s$ and fixed image $\bm{I}^s$, where $\bm{M}^s$ is the reconstructed moving image obtained from the previous scale $s/2$ downsampled to scale $s$ and $\bm{I}^s$ is the fixed image downsampled to scale $s$. At the coarsest scale $S$, $\bm{M}^S$ is taken to be downsampled moving image to scale $S$.  Let $\phi^s$ denote the registration field obtained at each scale $s$. (Note that $\phi^s$ covers the original spatial domain of the moving image, different from $\tilde\phi^s$, which is a residual mapping between down-sampled domains, i.e., the domain of $\bm{I}^s$.) Then the reconstructed image from scale $s/2$ is $\mathcal{T}(\bm{M}, \phi^{s/2})$ and $\bm{M}^s$ is the down-sampled version of this image to scale $s$.  By default, we assume $\phi^{S/2}$ is the identify map because $S$ is the coarsest scale, and $\mathcal{T}(\bm{M}, \phi^{S/2}) = \bm{M}$. 

More specifically, at scale $s$, we first resize the moving image $\mathcal{T}(\bm{M}, \phi^{s/2})$ and the fixed image $\bm{I}$ to the current scale $s$ for data preparation as
\begin{equation}\label{eq:nmsr}
\begin{aligned}
    \bm{M}^s = \mathcal{P}_1 ( \mathcal{T}({\bm{M}}, \phi^{s/2}), s ), \quad \bm{I}^s = \mathcal{P}_1 (\bm{I}, s) 
\end{aligned}
\end{equation}
where $\bm{M}^s$ is the downsampled reconstructed image and $\mathcal{P}_1$ is the down-sampling operator, ${\bm{M}}$ is the moving image. We employ a U-Net $\mathcal{N}(\cdot, \cdot; \bm{\theta}^s)$ to model the residual registration field of scale $s$, $\Tilde{\phi^s} = \mathcal{N}(\bm{I}^s, \bm{M}^s; \bm{\theta}^s)$, where $\bm{\theta}^s$ is the parameters in the U-Net.

\subsection{Multi-Scale Registration Field Aggregation}\label{sec:regfield_aggreg}
After the test-time training, we obtain the optimal parameters $\bm{\theta}^s_{\star}$ of the neural network and calculate the residual registration field $\Tilde{\bm{\phi}}^{s}$ for the last scale's reconstructed image $\mathcal{T}({\bm{M}}, {\bm{\phi}}^{s/2})$ of scale $s$. We calculate the registration field $\bm{\phi}^{s}$ for the moving image $\bm{M}$ by \textit{aggregating} registration field $\bm{\phi}^{s/2}$ of the last scale $s/2$ and intermediate/residual registration field $\Tilde{\bm{\phi}}^{s}$. 
For each pixel position $\bm{p}$ in the moving image $\bm{M}$, we obtain the final position $\hat{\bm{p}}$ and the combined registration field $\bm{\phi}^{s}$ by
\begin{equation}\label{eq:field_fusion}
\begin{aligned}
  \bm{\phi}^{s} &= \phi^{s/2} \circ \Tilde{\phi^{s}} \\
    \bm{\phi}^{s}(\bm{p}) &= \hat{\bm{p}} - \bm{p} \\&= \bm{p} + \bm{\phi}^{s/2}(\bm{p}) + \mathcal{P}_2 (\frac{1}{s} \Tilde{\phi}^{s}, \frac{1}{s})(\bm{p} + \bm{\phi}^{s/2}(\bm{p}) ) - \bm{p} \\
    & = \phi^{s/2}(\bm{p}) + \mathcal{P}_2 (\frac{1}{s} \Tilde{\phi}^{s}, \frac{1}{s})(\bm{p} + \phi^{s/2}(\bm{p}) ),
\end{aligned}
\end{equation}
where $\mathcal{P}_2$ is the \textit{linear interpolation} operator for up-sampling and we use linear intepolation to calculate the field for point $\bm{p} + \phi^{s/2}(\bm{p})$ in the current generated \textit{residual registration field} $\Tilde{\phi}^{s}$. We do not need to calculate the combined registration for the coarsest scale $S$ because $\phi^{S/2}$ is a zero field as predefined. 

\section{Results}\label{sec:result}
We validate test-time training for multi-scale deformable registration, including ablation study, on three datasets.  
\subsection{Data}\label{sec:data}
We employ a 3D hippocampus MRI dataset from the medical segmentation decathlon~\cite{simpson2019large} to validate the proposed method for registration-based segmentation. On the hippocampus dataset, we randomly split the dataset into 208 training images and 52 test images. There are two foreground categories, hippocampus head and hippocampus body. We re-sample the MR images to $1\times 1 \times 1$ $\mathrm{mm}^3$ spacing. To reduce the discrepancy of the intensity distributions of the MR images, we calculate the mean and standard deviation of each volume and clip each volume up to six standard deviations. Finally, we linearly transform each 3D image into range $[0, 1]$. The image size is within $48 \times 64 \times 48$ voxels.

We collect echocardiograms from 19 patients with 3,052 frames in total for myocardial tracking, and contrast echocardiograms from 71 patients with 11,462 frames in total for cardiac blood tracking. Contrast enhanced echocardiography-based vortex imaging has been used in patients with cardiomyopathy, LV hypertrophy, valvular diseases and heart failure. For testing, we randomly choose three patients' echocardiograms with 291 frames for myocardial tracking, and three patients' echocardiograms with 216 frames for cardiac blood tracking from the two datasets. The rest of echocardiograms are used for training. We use large training set to facilitate the learning based-method, VoxelMorph~\cite{balakrishnan2018unsupervised}, to perform well. The frame per second (FPS) of ultrasound for myocardial tracking is 75 and the FPS for cardiac blood tracking is from 72 to 93. Because of the consistency of FPS, it has little impact on the results. All echocardiograms have \textit{no registration field ground truth}.

We only use the first channel of echocardiography images (e.g., treated as gray-scale images) with the pixel values normalized to be in $[0, 1]$ by being divided by 255. The image size is $1024 \times 768$. To remove the background and improve the robustness of registration, we conduct registration on region of pixel value within the range $[0.005, 0.995]$ which is the default setting in ANTs~\cite{avants2008symmetric,avants2009advanced}.  To obtain a smooth boundary for stable optimization, we further employ morphological dilation with a disk-shaped structure of radius of 16 pixels to extract myocardial region.  For cardiac blood flow tracking, we extract cardiac blood region by 1) creating masks of the left ventricular blood pool at the end of the systole and the end of the diastole, 2) using active contour model to fit 100 uniformly sampled spline points along a circle into the boundary of cardiac blood mask~\cite{kass1988snakes}, 3) using linear interpolation to get 100 interpolated spline points for each frame, 4) using radial basis function in interpolation to obtain the final smooth cardiac blood boundary from the 100 spline points. Removing myocardial region is crucial to cardiac blood tracking.

\subsection{Performance Evaluation and Implementation}\label{sec:impl}  
For unsupervised learning, one of the main challenges is the model evaluation. Manually labeling the corresponding points for evaluation is time-consuming, laborious and inaccurate, because the image size is typically large especially for ultrasound images with low signal-to-noise ratio. We can use segmentation accuracy to evaluate registration by conducting registration-based segmentation, which employs the most similar training image based on NLCC as the moving image and obtains segmentation prediction by transforming the training label with the predicted registration field. For the segmentation, we compare our method with NiftyReg~\cite{modat2010fast}, ANTs~\cite{avants2008symmetric,avants2009advanced}, VoxelMorph~\cite{balakrishnan2018unsupervised} and NeurReg~\cite{zhu2019neurreg} which uses simulated registration field to conduct supervised learning. ANTs and NiftyReg are traditional optimization-based methods and VoxelMorph is a deep learning-based unsupervised registration with the same network structure and loss function as our method for a fair comparison. We use Dice coefficient (DSC) as the evaluation metric, defined to be $\frac{2 \times TP}{2 \times TP + FN + FP}$, where $TP$, $FN$, and $FP$ are true positives, false negatives, false positives, respectively. Because the image size is within $48 \times 64 \times 48$ voxels, we use a $5 \times 5 \times 5$-voxel window in the reconstruction loss $\mathcal{L}_{R}$ in Eq.~\eqref{eq:l_R}. We follow the same settings as NeurReg and the performances of the compared methods are reported from~\cite{zhu2019neurreg}.  

For the motion analysis, we use the last frame as the moving image and the current frame as the fixed image. Because there is \textit{no registration ground truth} for tissue tracking based on echocardiogram, we use reconstruction based metrics, i.e., the mean square error (MSE) and the normalized local cross correlation (NLCC) with radius of ten pixels to replace pixel position based evaluation metric. We calculate the average MSE and NLCC over all frame pairs with the linearly normalized pixel value of range $[0, 1]$. For MSE and NLCC of one pair of frames, we take the average of square error and normalized local cross correlation over the masked region obtained from Section~\ref{sec:data}. The method with low MSE and high NLCC has good reconstruction and is the preferred method. We compare our approach to ANTs~\cite{avants2008symmetric,avants2009advanced}, and VoxelMorph~\cite{balakrishnan2018unsupervised}. We use a $6 \times 6$-voxel window for reconstruction loss $\mathcal{L}_{R}$. For a fair comparison, we use the same network structure and number of channels in each convolution as VoxelMorph. 

For the purpose of ablation studies, we report the results of self-supervision optimized registration (SSR), self-supervision optimized multi-scale registration (SSMSR (1)) and different scales of self-supervision optimized multi-scale registration (SSMSR ($s$)). We use two scales $1/2$ and $1$ on hippocampus dataset and four different scales $1/8$, $1/4$, $1/2$ and $1$ on echocardiogram datasets because echocardiogram is relative large ($1024 \times 768$) and hippocampus image size is relative small ($48 \times 64 \times 48$). We use learning rate of $1\times10^{-3}$ and Adam optimizer to update the weights in neural networks for both our method and VoxelMorph~\cite{kingma2014adam}. The $\lambda$, which is the weight ratio of the smoothness loss and the reconstruction loss, is set to be $10$ based on the performance on the validation set. We set the number of optimization steps to $3,500$ per test pair or ultrasound sequence for the self-supervision optimized registration (of each scale), and set the number of iterations to $3,500 \times$ the number of training ultrasound sequences for VoxelMorph for fair comparison. To improve the generalization performance of our method, we also firstly update the weights on the training set the same as VoxelMorph.  In the experiment, we find, for self-supervision optimized multi-scale registration, the consecutive optimization across scales, i.e., weights from scale $s$ using optimized weights from scale $s/2$ as initialization, yields better registration. 

We conduct deformable registration and use the recommended hyper-parameters for ANTs and NiftyReg in~\cite{balakrishnan2018unsupervised}, because the field of view in the hippocampus and cardiac tissues is roughly aligned during image acquisition. For ANTs, we use three scales with 200 iterations each, B-Spline SyN step size of 0.25, updated field mesh size of five. For NiftyReg, we use three scales, the same local negative cross correlation objective function with control point grid spacing of five voxels and 500 iterations. We have tried our best to tune the best hyper-parameters for ANTs and VoxelMorph.

\subsection{Registration-Based Segmentation}\label{sec:exp_segm}
On the Hippocampus dataset, the image size is within $48 \times 64 \times 48$ voxels, which is small. We only use two scales for multi-scale registration. For ablative study, we calculate DSCs of test-time training, self-supervision optimized, registration without multi-scale scheme (SSR), self-supervision optimized multi-scale registration of scale 1/2 (SSMSR (1/2)) and  final self-supervision optimized multi-scale registration (SSMSR (1)) in Table~\ref{tab:seg_hippo}. We compare our method with previous registration approaches including ANTs, NiftyReg and recently proposed NeurReg~\cite{zhu2019neurreg}. We follow the same experimental settings as NeurReg. The results of ANTs, NiftyReg, VoxelMorph and NeurReg are reported in~\cite{zhu2019neurreg}. We use only one atlas for each test image based on NLCC similarity score from VoxelMorph. The segmentation comparisons are listed in Table~\ref{tab:seg_hippo}. With the same set of atlases, self-supervision optimized registration yields better segmentation than VoxelMorph, which means that the alleviation of domain shift by self-supervised optimization improves the accuracy of registration-based segmentation. Test-time training by self-supervision optimized multi-scale registration further improves registration-based segmentation accuracy based on segmentation dice score on both classes.
\begin{table}[t]
\begin{center}
\caption{Segmentation (Dice scores, \%) on hippocampus dataset.} 
\label{tab:seg_hippo}
\begin{tabular}{lll}
\hline\noalign{\smallskip}
Method & hippocampus head/body & Average $\uparrow$\\
\noalign{\smallskip}
\hline
\noalign{\smallskip}
ANTs &80.86$\pm$5.13/78.34$\pm$5.24 & 79.60 \\ 
NiftyReg &80.53$\pm$4.86/77.92$\pm$5.47 & 79.23 \\ 
VoxelMorph &\begin{tabular}{@{}c@{}}78.92$\pm$6.79/75.85$\pm$8.13\end{tabular} & 77.39\\ 
NeurReg &\begin{tabular}{@{}c@{}}78.99$\pm$6.24/78.72$\pm$7.47\end{tabular} & 78.86 \\ 
\hline
\begin{tabular}{@{}c@{}}SSR (w/o multi-scale)\end{tabular}  &\begin{tabular}{@{}c@{}}80.61$\pm$5.76/78.73$\pm$8.14\end{tabular}  &  79.67\\ 
\begin{tabular}{@{}c@{}}SSMSR (1/2) (coarsest scale)\end{tabular}  &\begin{tabular}{@{}c@{}}77.66$\pm$4.90/75.08$\pm$8.01\end{tabular}  & 76.37 \\ 

SSMSR (1) &\begin{tabular}{@{}c@{}}\bf{81.18$\pm$5.06/79.13$\pm$7.04}\end{tabular} & \bf{80.16} \\
\hline
\end{tabular}
\end{center}
\end{table}

\subsection{Registration-Based Tissue Dense Tracking}\label{sec:exp_ssr}  
To validate the effectiveness of our method on noisy and large deformation ultrasound images, we calculate the performances of registration fields from ANTs, VoxelMorph, test-time training by self-supervision optimized registration (SSR) and self-supervision optimized multi-scale registrations (SSMSR ($s$)).  Quantitative comparison results of these models on both myocardial and cardiac blood flow dense tracking are shown in Table~\ref{tab:comp}. 

\begin{table}[t]
\begin{center}
\caption{Comparisons on myocardial (Upper) and cardiac blood flow (Lower) dense tracking among ANTs, VoxelMorph and Ours. }\label{tab:comp} 
\begin{tabular}{lll}
\hline
Methods & MSE (10\textsuperscript{-3}) $\downarrow$      & NLCC (10\textsuperscript{-1})  $\uparrow$  \\
\hline
ANTs & 15.5179$\pm$9.4637 & 3.1597$\pm$1.3913 \\
VoxelMorph & 1.2266$\pm$0.5457 & 4.7363$\pm$0.5457  \\
\hline
SSR (w/o multi-scale) & 1.2158$\pm$0.5473 & 4.7809$\pm$0.4952  \\  
SSMSR (1/8) (coarest scale) & 1.6937$\pm$0.5616 & 4.2084$\pm$0.5086 \\  
SSMSR (1/4) & 1.3504$\pm$0.4885 & 4.4140$\pm$0.5005   \\  
SSMSR (1/2) & 1.0929$\pm$0.3775 & 4.7155$\pm$0.4569  \\  
SSMSR (1) & \bf{0.9206$\pm$0.3236} & \bf{5.0881$\pm$0.4107}  \\  \hline \hline

ANTs & 3.9344$\pm$1.4660 & 4.2062$\pm$1.0576 \\
VoxelMorph  & 5.8654$\pm$1.6943 & 3.3462$\pm$0.5891 \\
\hline
SSR (w/o multi-scale)  & 5.7780$\pm$1.6222 & 3.3942$\pm$0.5868 \\  
SSMSR (1/8) (coarest scale)  & 6.8108$\pm$1.8189 & 2.3948$\pm$0.5495 \\  
SSMSR (1/4)  & 5.6106$\pm$1.2424 & 2.7331$\pm$0.6779  \\  
SSMSR (1/2) & 4.5089$\pm$1.0272 & 3.3879$\pm$0.7524 \\  
SSMSR (1) & \bf{3.7944$\pm$0.9384} & \bf{4.2519$\pm$0.7181} \\  \hline

\end{tabular}
\end{center}
\end{table}
From Table~\ref{tab:comp}, we highlight the following observations. 1) SSMSR (1) achieves the best performances, and outperforms ANTs in terms of both MSE and NLCC on both myocardial and cardiac blood tracking, likely due to the representation and optimization efficiency of deep neural nets; Deep learning has a good capacity to model the registration field. 2) SSR yields consistently better results than VoxelMorph, demonstrating the efficacy of self-supervised optimization during the test phase for improving registration field estimation and reducing the estimation gap between training and testing. 3) SSMSR (1) obtains better performance than SSR on all experiments, demonstrating the benefit of sequential multi-scale optimization in echocardiogram registration. The multi-scale scheme alleviates the over-optimization of reconstruction loss compared with ANTs, which can be visually noticed from Fig.~\ref{fig:muscle} and \ref{fig:blood} in Section~\ref{sec:visual}. The multi-scale registration with consecutive test-time training, SSMSR (1), achieves the best performance on the two tasks and outperforms traditional registration, ANTs, and current unsupervised registration, VoxelMorph, on the two tasks based on the two metrics. 

\section{Discussion}\label{sec:dis}
\subsection{Understand Test-Time Training}\label{sec:under_sso}
\begin{figure}[t]
	\centering
		\begin{minipage}{0.9\linewidth}
			\includegraphics[width=\textwidth,trim=336 192 304 128,clip]{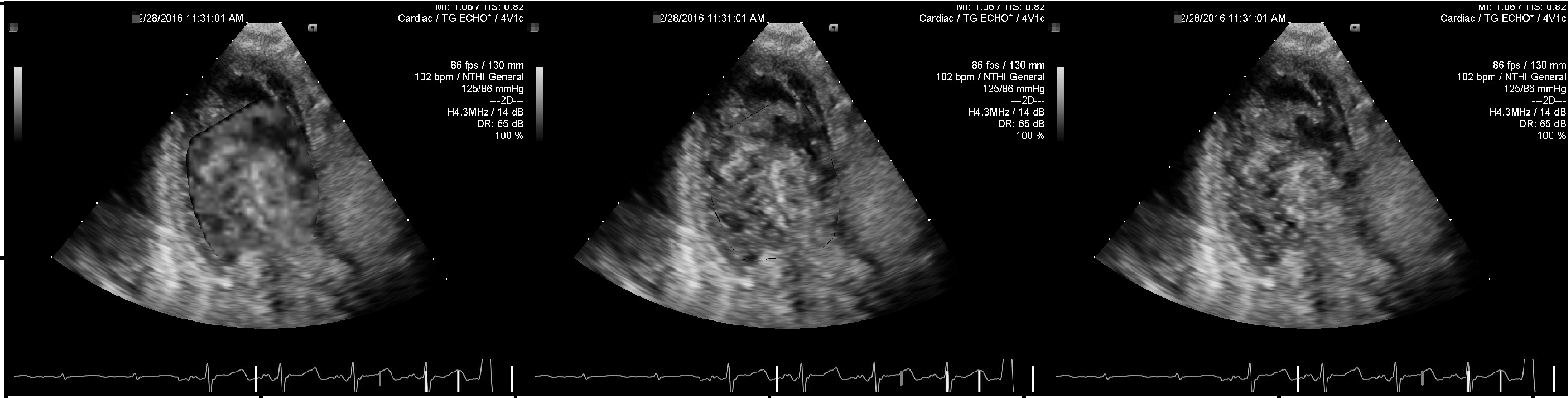}
		\end{minipage}
		\begin{minipage}{0.9\linewidth}
			\includegraphics[width=\textwidth,trim=42 24 38 16,clip]{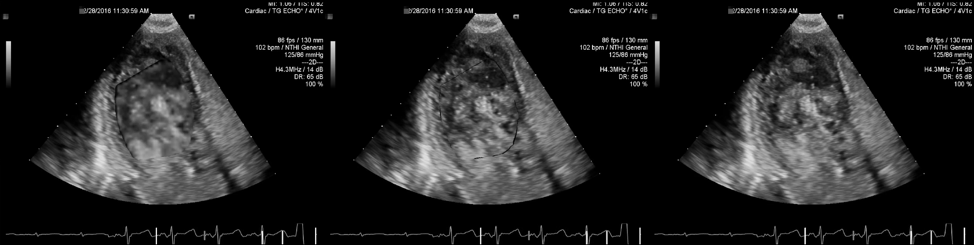}
		\end{minipage}
	\caption{Reconstruction comparison for cardiac blood from test-time training based multi-scale registration (scale 1/8) w/o (first), i.e., VoxelMorph~\cite{balakrishnan2018unsupervised}, w/ (middle) test-time training and ground truth image (last column).} 
	\label{fig:vis_ssr}
\end{figure}

To further understand the importance of test-time training in each scale of multi-scale registration, we visualize the reconstructions of cardiac blood from test-time training based multi-scale registration (scale 1/8) with (second column) and without (first column) test-time training respectively in Fig.~\ref{fig:vis_ssr}. From the visual comparison, multi-scale registration without test-time training (first column) reconstructs worse blood details than multi-scale registration with test-time training (second column). In the multi-scale registration framework, the current registration relies on the previous reconstructed image which makes the optimization more difficult if the reconstruction from coarse level cannot do well.  
\subsection{Understand Multi-Scale Registration}\label{sec:visual} 
To further understand the self-supervision optimized multi-scale registration (SSMSR) in the test-time, we randomly choose two neighbor frames from ultrasound sequences and visualize the myocardial tracking results based on ANTs, VoxelMorph and the intermediate registration fields from SSMSR with four different scales in Fig.~\ref{fig:muscle}. More visualization results can be found in the supplemental materials.

\begin{figure}[t]
	\centering
		\begin{minipage}{0.3\linewidth}
			\includegraphics[width=\textwidth]{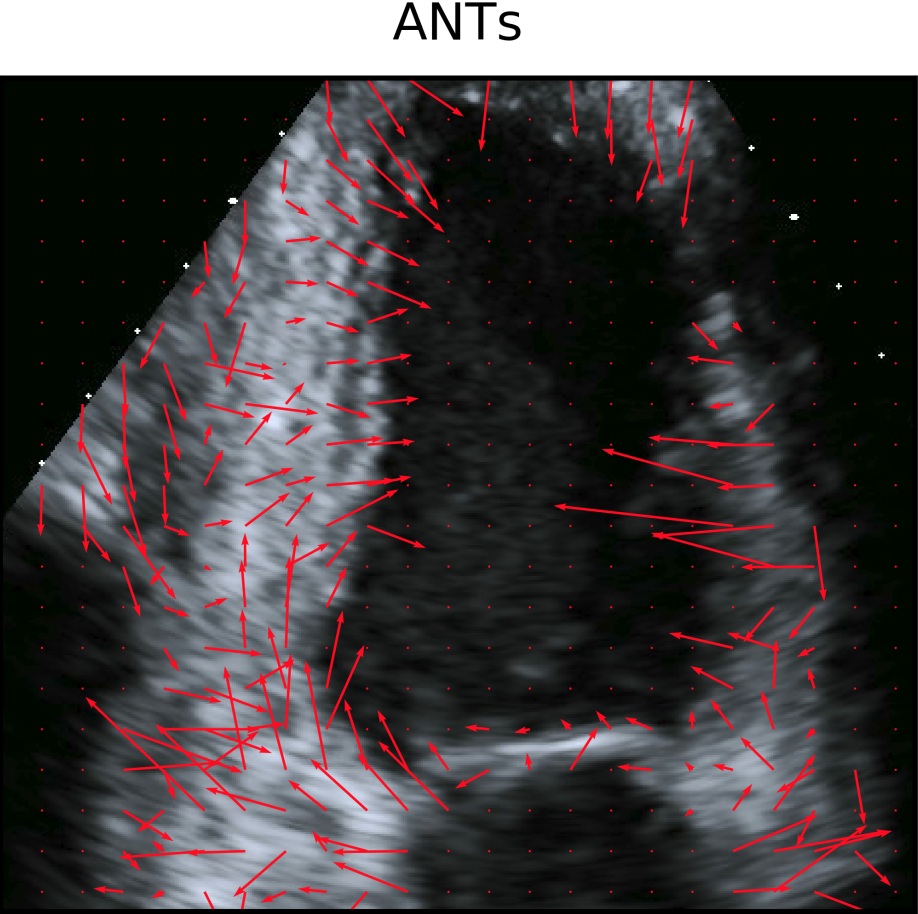}
		\end{minipage}
		\begin{minipage}{0.3\linewidth}
			\includegraphics[width=\textwidth]{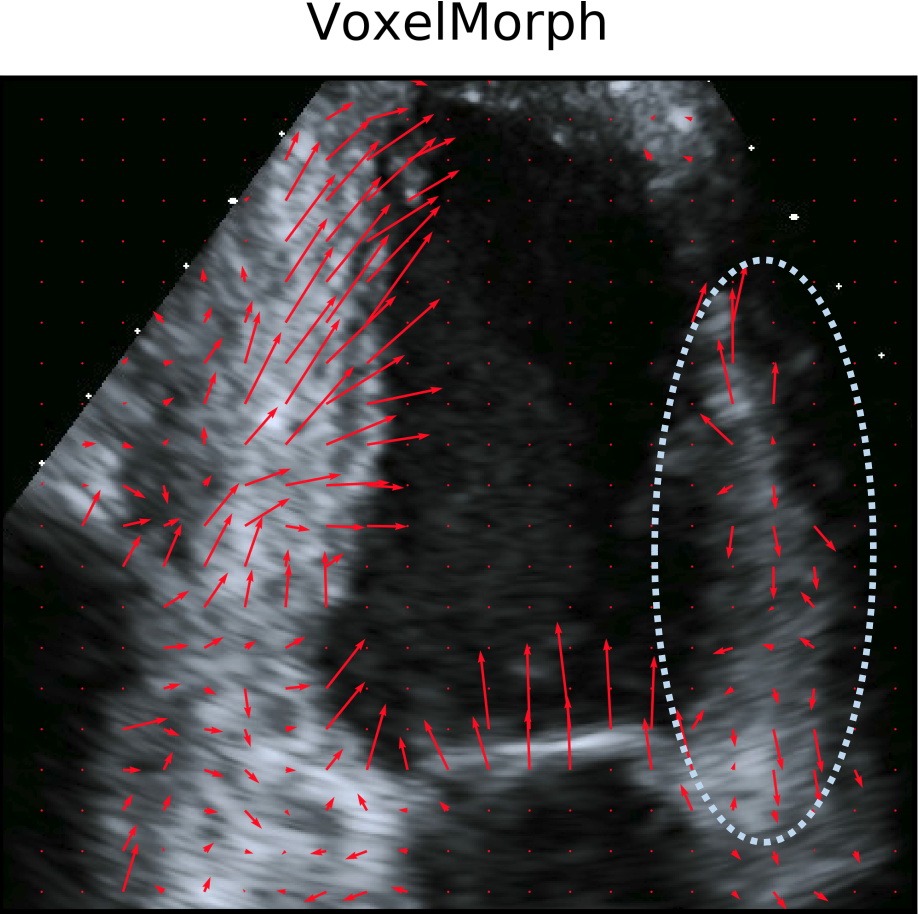}
		\end{minipage}
		\begin{minipage}{0.3\linewidth}
			\includegraphics[width=\textwidth,trim=0 0 0 12,clip]{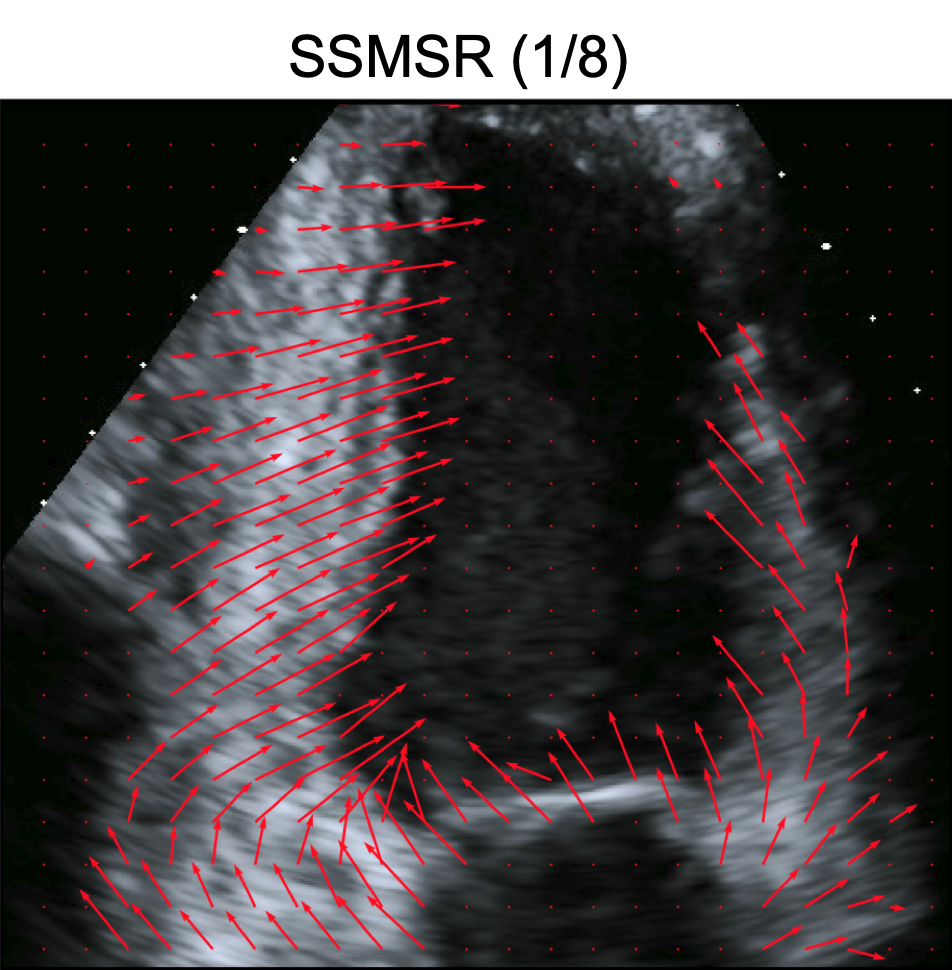} 
		\end{minipage} \\
		
		\begin{minipage}{0.3\linewidth}
			\includegraphics[width=\textwidth,trim=0 0 0 12,clip]{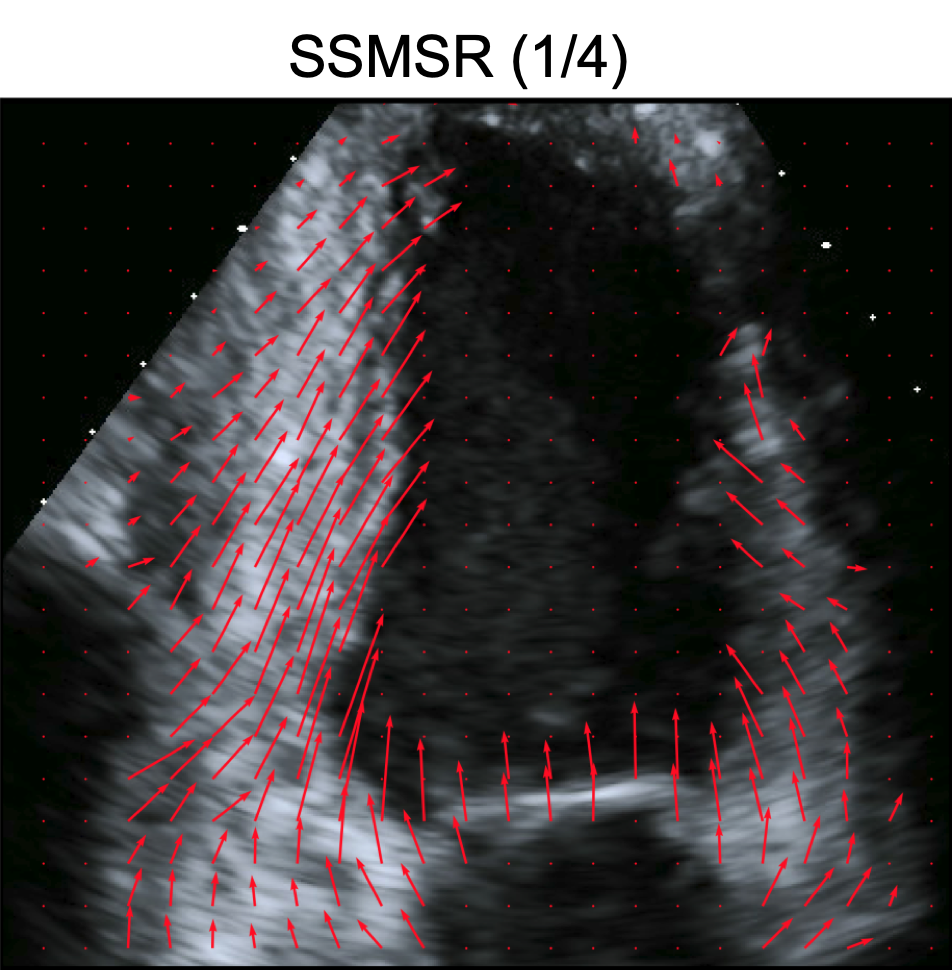}
		\end{minipage}
		\begin{minipage}{0.3\linewidth}
			\includegraphics[width=\textwidth,trim=0 0 0 12,clip]{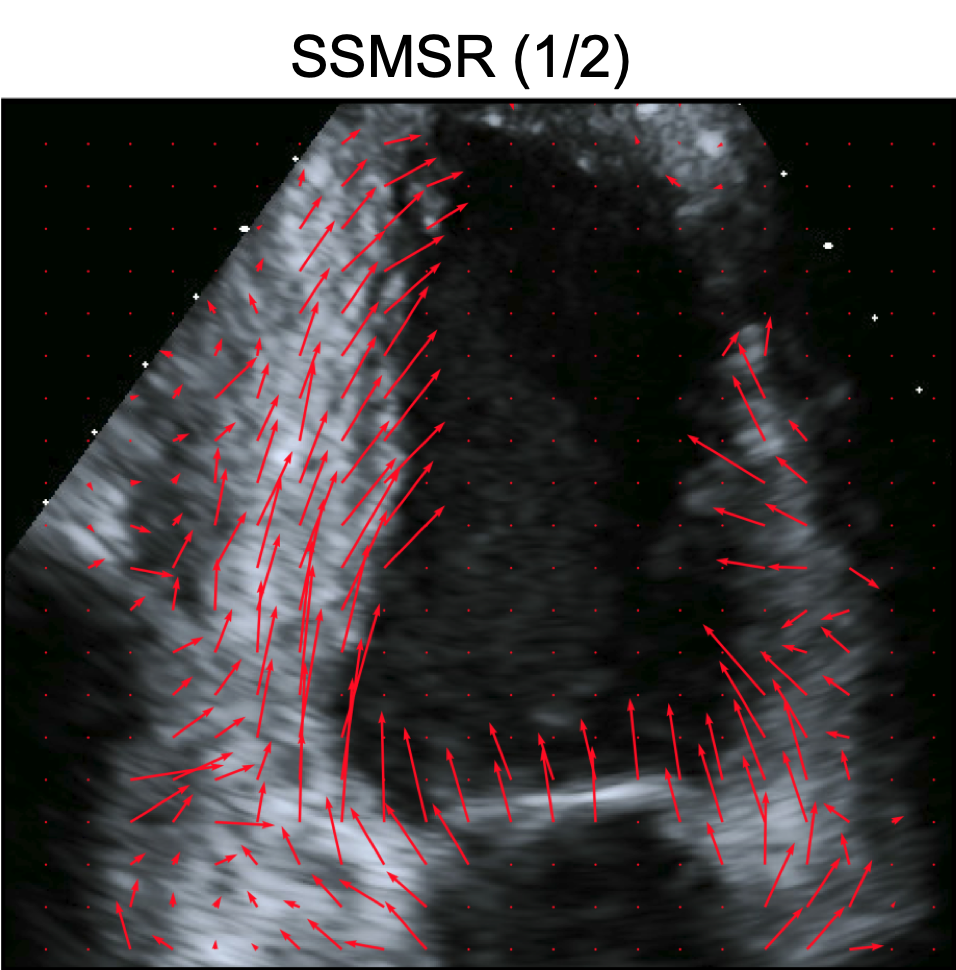}
		\end{minipage}
		\begin{minipage}{0.3\linewidth}
			\includegraphics[width=\textwidth,trim=0 0 0 12,clip]{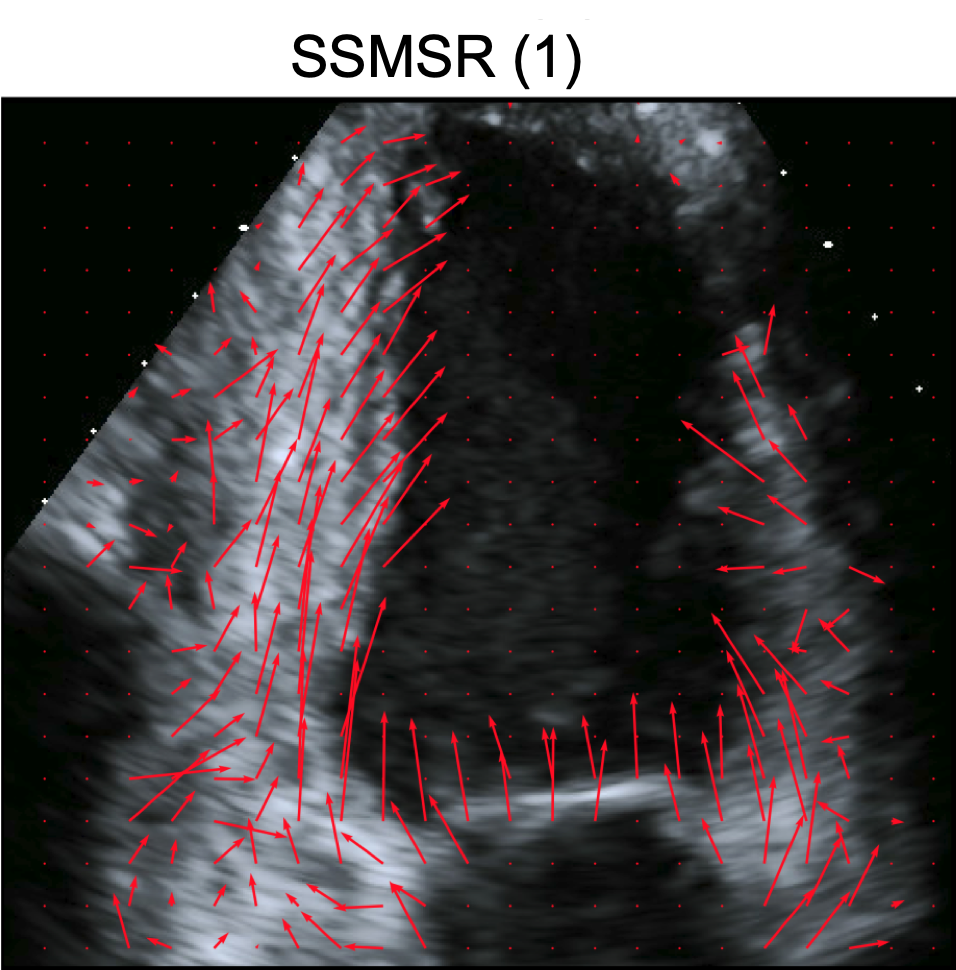}
		\end{minipage}
	\caption{Visualization of myocardial tracking based on ANTs, VoxelMorph~\cite{balakrishnan2018unsupervised} and Ours.} 
	\label{fig:muscle}
\end{figure}
From Fig.~\ref{fig:muscle}, we note that the registration field from ANTs is noisy, and the velocity direction from VoxelMorph for the right myocardial is incorrect because of contradictory in the estimated direction of myocardial. By contrast, SSMSR (1/8) produces the smoothest registration field, and SSMSR (1) generates more detailed velocity estimation that preserves both large and low-scale velocity variations.  The coarse-to-fine results illustrate the multi-scale optimization scheme coupled with deep neural nets can be very effective in dealing with the highly challenging case of image registration in echocardiograms.

\begin{figure}[t]
	\centering
		\begin{minipage}{0.3\linewidth}
			\includegraphics[width=\textwidth]{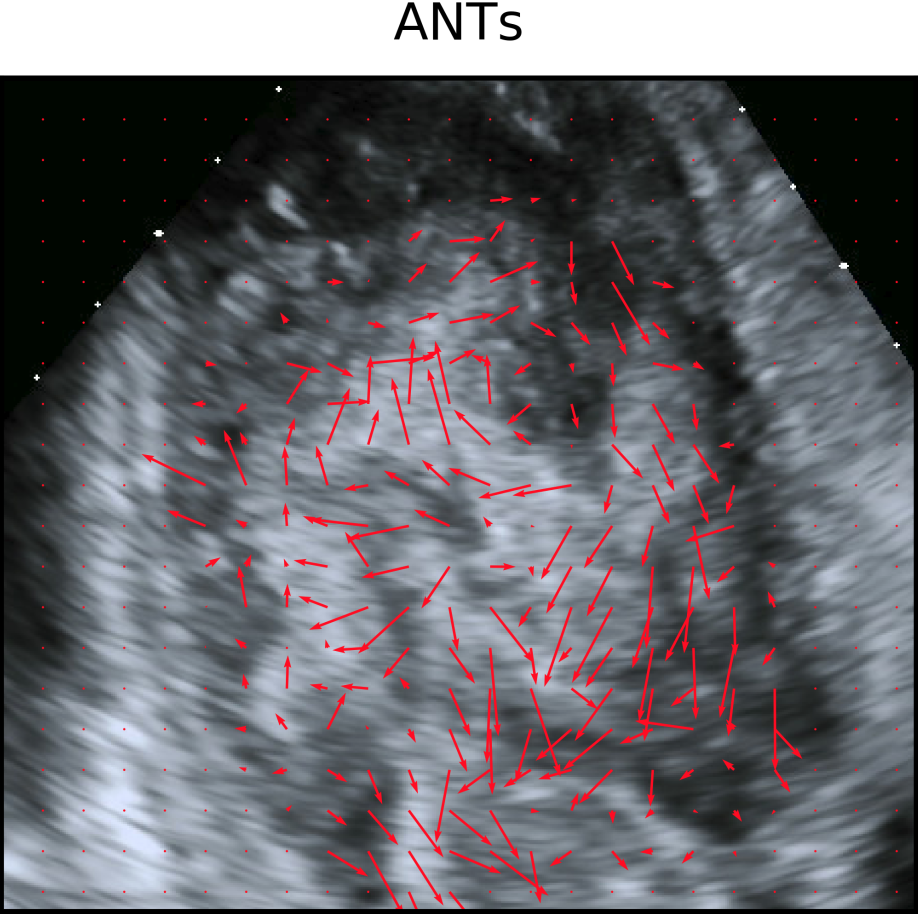}
		\end{minipage}
		\begin{minipage}{0.3\linewidth}
			\includegraphics[width=\textwidth]{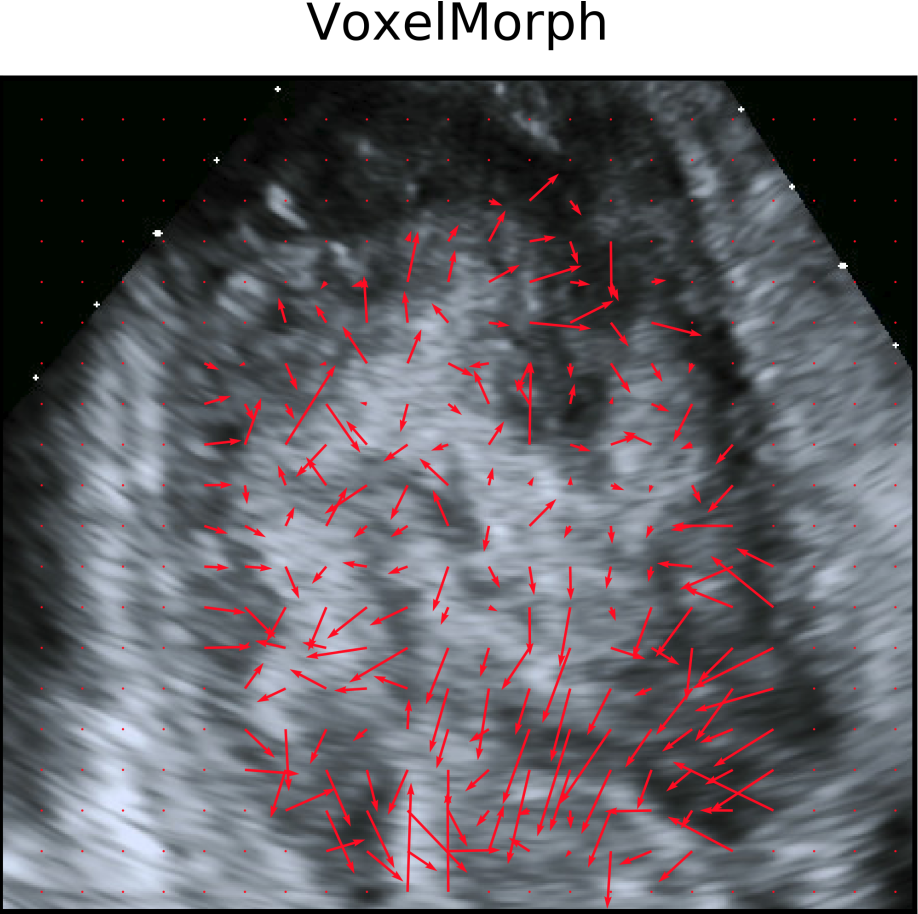}
		\end{minipage}
		\begin{minipage}{0.3\linewidth}
			\includegraphics[width=\textwidth,trim=0 0 0 12,clip]{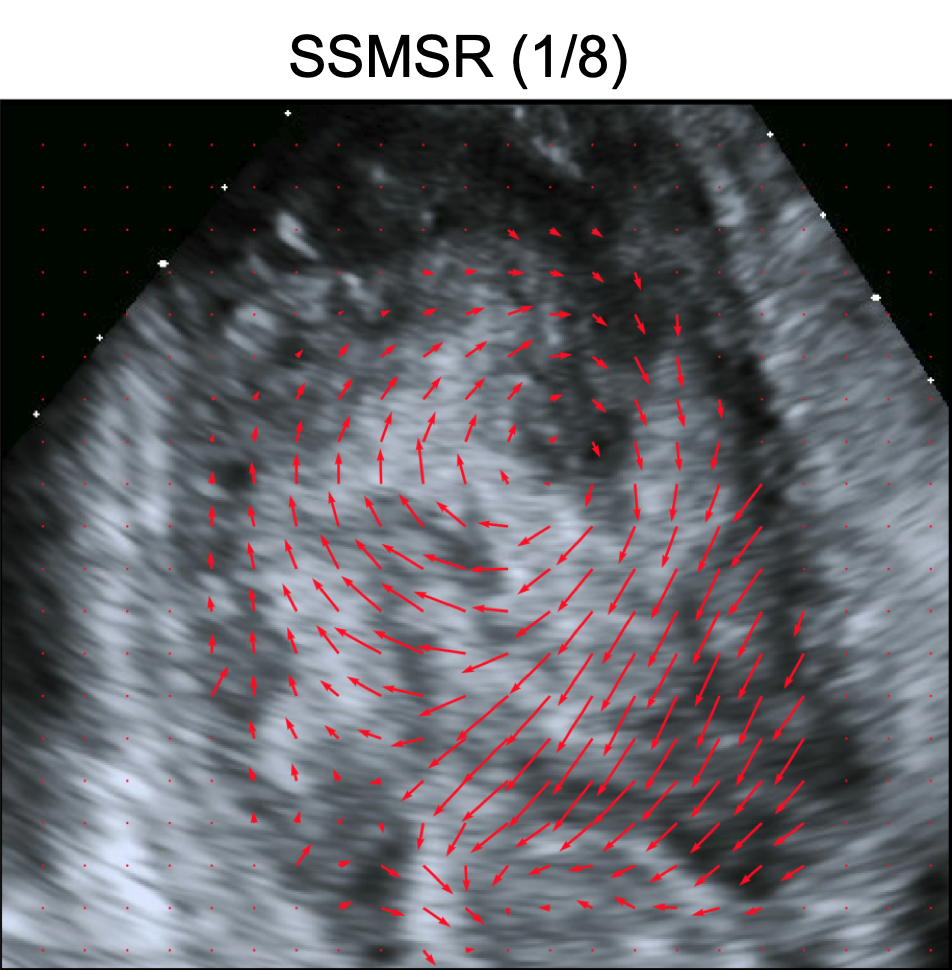}
		\end{minipage} \\
		
		\begin{minipage}{0.3\linewidth}
			\includegraphics[width=\textwidth,trim=0 0 0 12,clip]{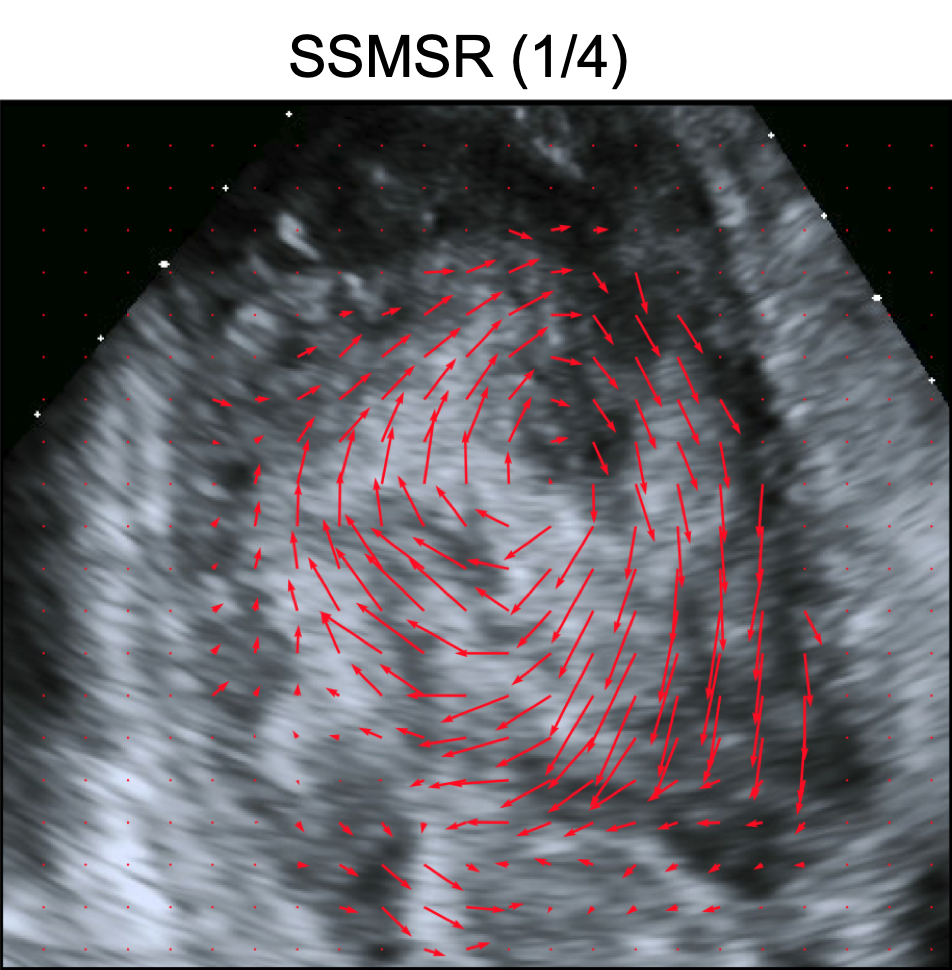}
		\end{minipage}
		\begin{minipage}{0.3\linewidth}
			\includegraphics[width=\textwidth,trim=0 0 0 12,clip]{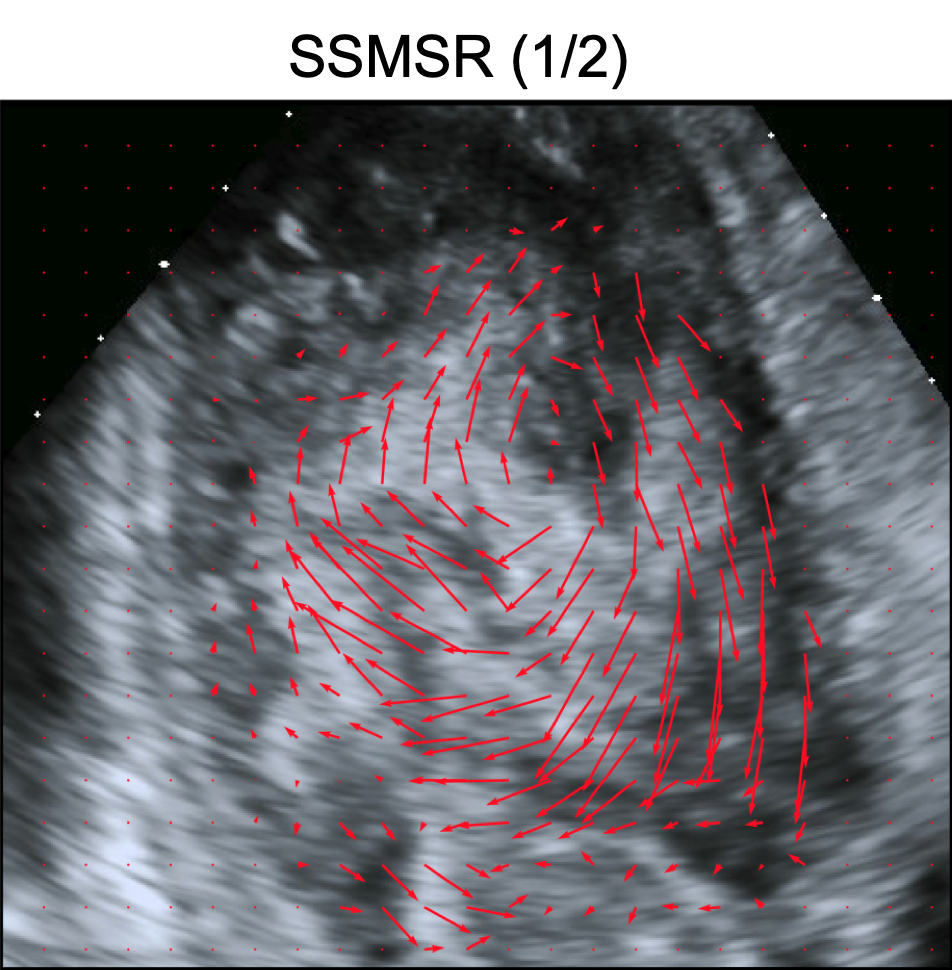}
		\end{minipage}
		\begin{minipage}{0.3\linewidth}
			\includegraphics[width=\textwidth,trim=0 0 0 12,clip]{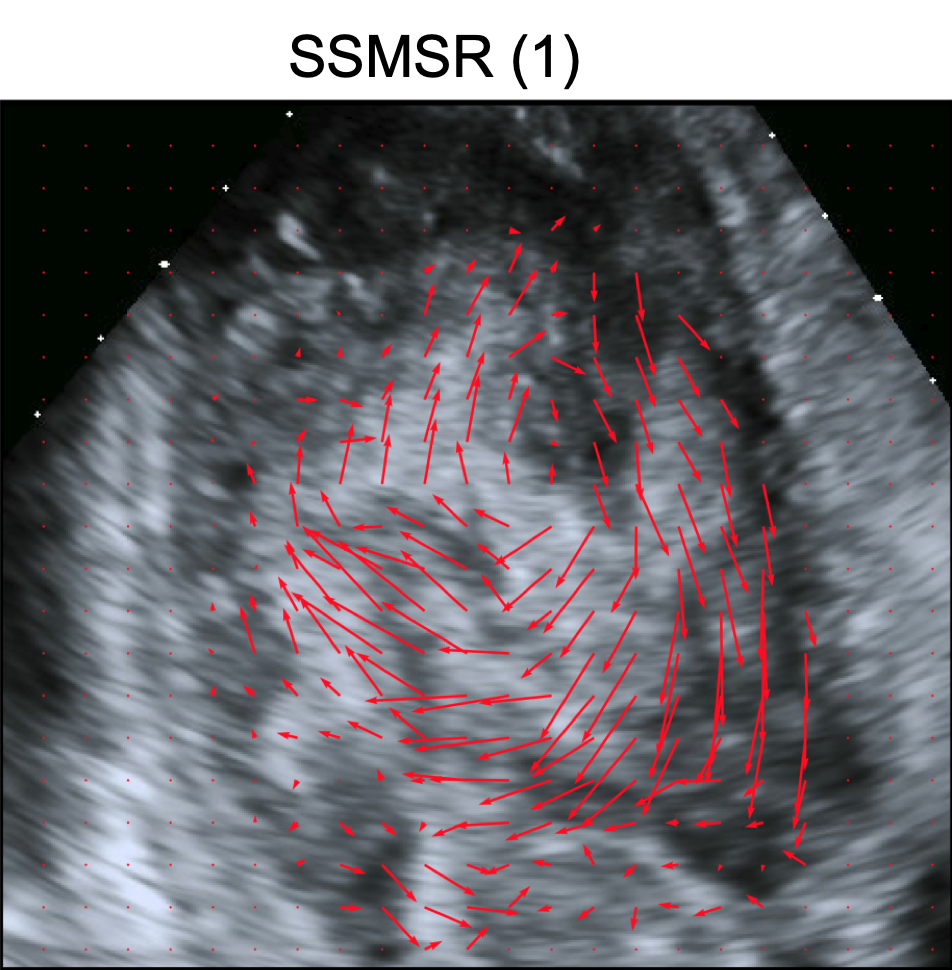}
		\end{minipage}
	\caption{Visualization of cardiac blood flow tracking based on ANTs, VoxelMorph~\cite{balakrishnan2018unsupervised} and Ours.}  
	\label{fig:blood}
\end{figure}

To facilitate the understanding of proposed SSMSR in the test-time on dense blood flow tracking based on echocardiograms, we also visualize the cardiac blood flow tracking results from ANTs, VoxelMorph and the intermediate registration fields from SSMSR with four different scales in Fig.~\ref{fig:blood}. We randomly choose two neighbor frames from these ultrasound images.  

From the registration fields generated from ANTs and VoxelMorph in Fig.~\ref{fig:blood}, we cannot easily recognize the vortex in cardiac blood flow. By contrast, the vortex flow pattern from SSMSR is readily recognizable. The general vortex pattern is apparent from the coarsest level registration by SSMSR (1/8), followed by finer-scale registrations to introduce details of local velocity field variations. The final velocity field produced by SSMSR (1) includes both easily recognizable vortex flow, as well as details of local field variations. 

\subsection{Computational Cost}\label{sec:comp}
We compare the computation cost on the echocardiogram dataset. NiftyReg takes the same scale of time as ANTs as validated in~\cite{zhu2019neurreg}. For ANTs, we cannot find implementation on GPU and the average computational time is 214.10$\pm$54.04 seconds for the registration of two consecutive frames on 12 processors of Intel i7-6850K CPU @ 3.60GHz. For an ultrasound sequence of 50 frames, the computational time is about {\bf{three hours}} for ANTs. Because the inference of VoxelMorph only relies on one feed forward pass of deep neural network, the average computational time is 0.11$\pm$0.47 seconds for one pair frames on one NVIDIA 1080 Ti GPU. The test-time training based multi-scale registration takes 279.97, 101.65, 68.79, 66.09 seconds for self-supervision optimization with the scale 1, 1/2, 1/4, 1/8 respectively on one ultrasound sequence of 49 frames by one NVIDIA 1080 Ti GPU. The SSMSR takes less than {\bf{nine minutes}} in test time in total for one ultrasound sequence, achieving {\bf{20}} times speedup over ANTs even using four scales.

\section{Conclusions}\label{sec:conclu}
In this work, we propose a novel framework, test-time training based multi-scale registration, as a general framework for image deformable registration. To produce accurate registration field estimation from noisy medical images and reduce the estimation gap between training and testing, we incorporate test-time training in the registration framework. To handle large variations of registration fields, a multi-scale scheme is integrated into the proposed framework to reduce the over-optimization of similarity functions and provides a sequential residual optimization pathway which alleviates the optimization difficulties in registration. Our proposed method consistently outperforms previous approaches on both registration-based segmentation task of 3D MR images and myocardial and cardiac blood flow dense tracking task of echocardiograms.

In future work, the current framework can be extended by: 1) forcing the network to generate consistent displacement fields between moving and fixed images~\cite{godard2017unsupervised}, 2) adapting iterative module and improving the efficiency by predicting the final registration field directly~\cite{hur2019iterative}, 3) integrating registration into segmentation facilitated by the smart strategy for large GPU memory consumption~\cite{zhu2019anatomynet,zhu2020lamp,zhu2019neurreg}.  

{\small
\bibliographystyle{./bibliography/IEEEtran}
\bibliography{icra}}

\end{document}